\pdfoutput=1

\documentclass[11pt]{article}

\PassOptionsToPackage{table}{xcolor}
\usepackage{xcolor}        
\usepackage[final]{acl}

\usepackage[T1]{fontenc}           
\usepackage[utf8]{inputenc}        

\usepackage{times}                 

\usepackage{graphicx}              

\usepackage{booktabs}              
\usepackage{multirow}              
\usepackage{pifont}                

\usepackage{amsmath,amssymb,amsthm} 
\usepackage{latexsym}              
\usepackage{subfigure}             
\usepackage{subcaption}            
\usepackage{float}                 

\usepackage{microtype}             
\usepackage{inconsolata}           

\usepackage{algorithm}
\usepackage{algorithmic}

\usepackage{setspace}              
\usepackage[normalem]{ulem}        
\useunder{\uline}{\ul}{}           

\usepackage{tikz}                  
\usepackage{pgf-pie}               
\usepackage{bbding}                
\usepackage{wasysym}               
\usepackage{paralist}              

\title{HapticLLaMA \includegraphics[width=1.35em]{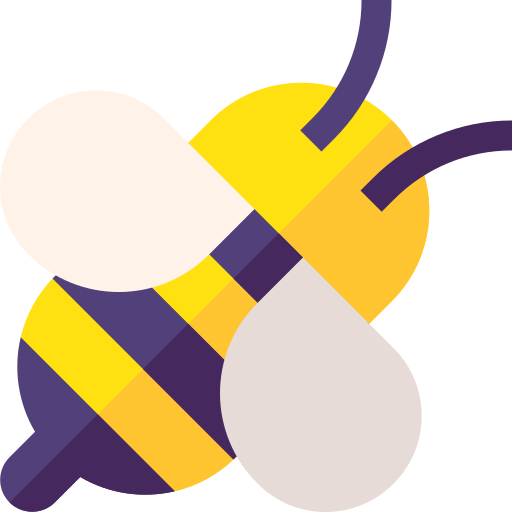}: A Multimodal Sensory \\Language Model for Haptic Captioning}

\author{Guimin Hu$^{1}$, Daniel Hershcovich$^{1}$, Hasti Seifi$^{2}$  \\
  $^1$University of Copenhagen \\
  $^2$Arizona State University \\
\texttt{rice.hu.x@gmail.com} \quad \texttt{dh@di.ku.dk} \quad \texttt{hasti.seifi@asu.edu}}

\begin{document}

\maketitle
\begin{abstract}
Haptic captioning is the task of generating natural language descriptions from haptic signals, such as vibrations, for use in virtual reality, accessibility, and rehabilitation applications. 
While previous multimodal research has focused primarily on vision and audio, haptic signals for the sense of touch remain underexplored. To address this gap, we formalize the haptic captioning task and propose HapticLLaMA, a multimodal sensory language model that interprets vibration signals into descriptions in a given sensory, emotional, or associative category. We investigate two types of haptic tokenizers, a frequency-based tokenizer and an EnCodec-based tokenizer, that convert haptic signals into sequences of discrete units, enabling their integration with the LLaMA model. HapticLLaMA is trained in two stages: (1) supervised fine-tuning using the LLaMA architecture with LoRA-based adaptation, and (2) fine-tuning via reinforcement learning from human feedback (RLHF). We assess HapticLLaMA’s captioning performance using both automated n-gram metrics and human evaluation.  
HapticLLaMA demonstrates strong capability in interpreting haptic vibration signals, achieving a METEOR score of 59.98 and a BLEU-4 score of 32.06 respectively. Additionally, over 61\% of the generated captions received human ratings above 3.5 on a 7-point scale, with RLHF yielding a 10\% improvement in the overall rating distribution, indicating stronger alignment with human haptic perception. These findings highlight the potential of large language models to process and adapt to sensory data.


\end{abstract}

\section{Introduction}
Humans perceive their environment through five primary senses: vision, hearing, touch (haptics), taste, and smell. Integrating these sensory modalities can enhance AI systems' ability to interpret human perception and behavior by providing a richer and more nuanced understanding of user context and sensory experiences. Haptic signals convey diverse information through tactile sensations, including physical attributes such as surface texture \cite{culbertson2014one}, emotional states like urgency or pleasantness \cite{yoo2015emotional}, and recognizable real-world cues such as heartbeats or buzzing of a bee \cite{seifi2015vibviz,seifi2017exploiting}. These tactile interactions have broad applications in areas including user interactions in virtual reality (VR), physical rehabilitation, blind user navigation, and gaming \cite{choi_augmenting_2021,seim_design_2022,katzschmann2018safe,yun_generating_2023,maclean_multisensory_2017}.


\begin{figure}[!t]
\centerline{\includegraphics[width=\linewidth]{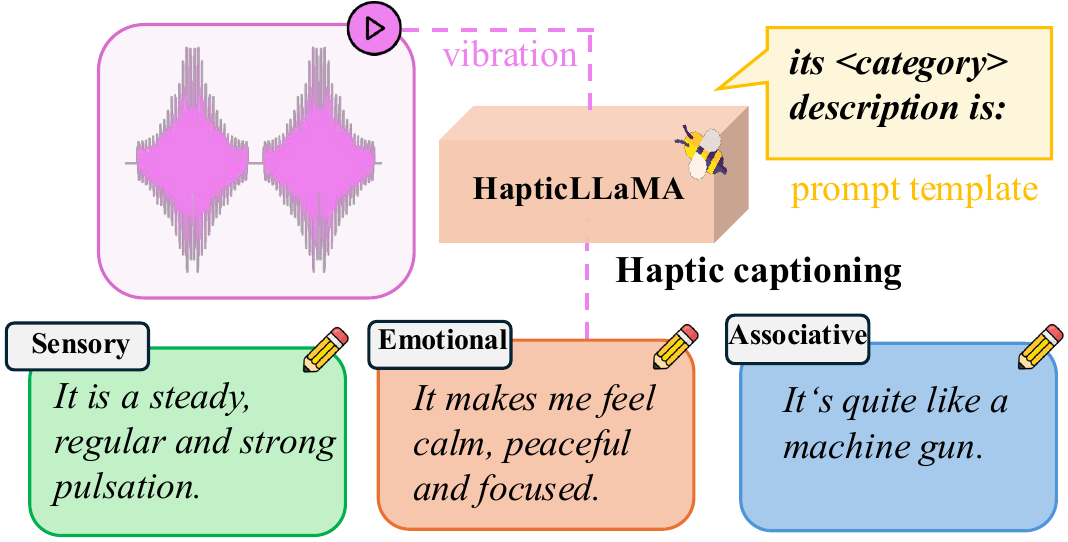}}
\caption{HapticLLaMA 
can generate sensory, emotional, and associative captions for an input vibration haptic signal with a tailored prompt template.}
\label{fig:teaser}
\end{figure}

Multimodal Large Language Models (MLLMs) have significantly advanced captioning across various modalities such as images~\cite{vinyals2015show,mokady2021clipcap}, video~\cite{iashin2020multi,wu2023cap4video}, and audio~\cite{zhang2022caption,liu2022visually}. \emph{Haptic captioning} involves generating textual descriptions (captions) of haptic signals, capturing sensory, emotional, and associative qualities (see Figure \ref{fig:teaser}). 
This process enhances the interpretability of physical signals by describing human perception of vibration patterns, similar to how vision-language models capture human perception of images. 
Additionally, haptic captioning offers a novel opportunity to develop datasets and benchmarks that probe the limits of AI capabilities, challenging models to accurately interpret and articulate complex physical sensations through natural language.

Unlike image and audio captioning, haptic captioning remains an underexplored domain. 
As a nascent field, haptic-language understanding presents unique challenges but also significant potential for advancing multimodal AI. The task involves two key challenges:
(1) the lack of established tokenization methods for representing haptic signals in a format suitable for language models; and (2) the absence of sensory multimodal models capable of processing haptic vibrations. To date, little work has investigated the ability of large language models (LLMs) to interpret haptic signals. 

To support the haptic captioning task, we introduce \textbf{HapticLLaMA}, the first multimodal haptic language model trained with two distinct haptic tokenizers that integrates haptic vibration signals and textual descriptions within a single framework.
Raw haptic signals are continuous time-series data, incompatible with the discrete token-based input required by LLMs. To address this,  we investigate two types of haptic tokenizers: a frequency-based tokenizer and an EnCodec-based tokenizer \cite{DBLP:journals/tmlr/DefossezCSA23}, both designed to convert vibrations into interpretable token sequences suitable for LLMs. 
HapticLLaMA is trained in two stages: (1) supervised fine-tuning using a LoRA-adapted \cite{hu2022lora} LLaMA architecture \cite{DBLP:journals/corr/abs-2302-13971}, and 
(2) reinforcement learning from human feedback \cite[RLHF;][]{ouyang2022training} applied using generated haptic captions rated by humans. 

Specifically, we contribute: 
(1) HapticLLaMA, the first haptic language model 
capable of generating sensory, emotional, and associative captions for vibration signals, 
(2) two haptic tokenizers that convert raw vibration inputs into sequences of discrete tokens,
(3) the VibRate dataset, containing 16,896 user-rated <vibration, caption, rating> samples, used to further incorporate human perception into the model via RLHF, and
(4) extensive experiments using automated n-gram metrics and human evaluation demonstrating that HapticLLaMA exhibits strong perceptual capabilities in describing vibration signals, achieving a METEOR score of 59.98 and a BLEU-4 score of 32.06, and receiving human ratings above 3.5 (on a 7-point scale) for over 61\% of the generated captions.

\begin{figure*}[!t]
\centerline{\includegraphics[width=0.96\textwidth]{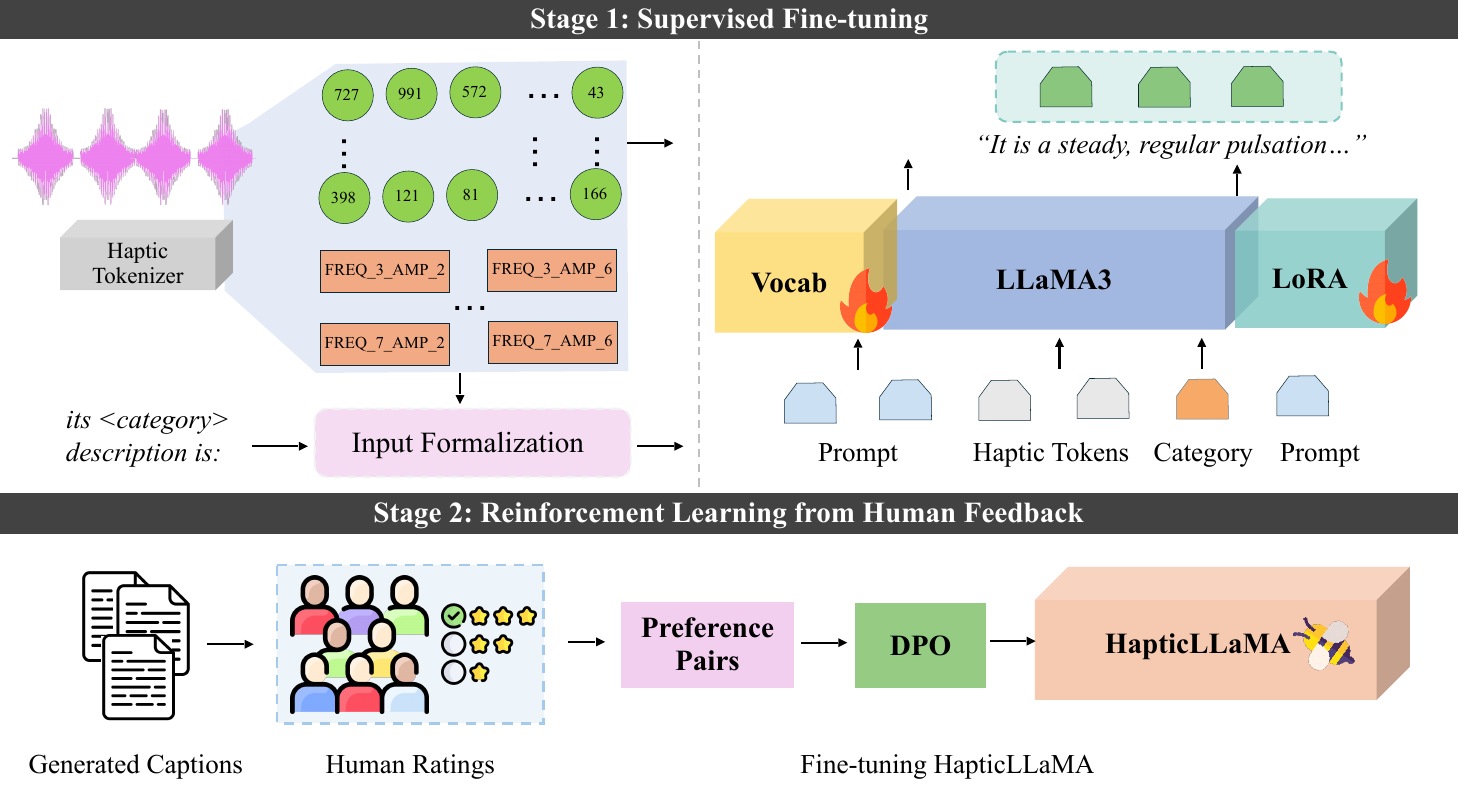}}
\caption{Overview of the two-stage process for constructing HapticLLaMA: (1) supervised fine-tuning, and (2) reinforcement learning from human feedback (RLHF).}
\label{fig:architecture}
\end{figure*}

\section{Related Work}

\subsection{Haptic Modality and Touch Datasets}

Early research on haptic language understanding ~\cite{obrist_talking_2013,seifi2015vibviz,seifi2017exploiting,dalsgaard_user-derived_2022} emerged within the human-computer interaction (HCI) domain, primarily through qualitative studies. While these works underscored the importance of understanding how users describe haptic experiences, they were limited in scale, typically focusing on fewer than 20 signals, and relied on manual analysis methods. Recent work proposed large-scale touch-language datasets for robotic sensing. Datasets such as TVL~\cite{fu2024touch}, TLV~\cite{cheng2024towards}, and Touch100k~\cite{cheng2024touch100k} use tactile images captured by deformable RGB-like sensors~\cite{yuan2017gelsight}. Others collected camera and touch sensor data from human interactions with objects~\cite{balasubramanian2024sens3,yang2022touch}. These datasets convey object properties like shape, size, and texture. While useful for training robots to perceive physical objects, these datasets lack programmable haptic feedback like vibrations from phones and VR controllers, which are common in user-facing applications. 
Recently, \citet{hu2024grounding} developed a pipeline for mapping emotional tags to haptic features, but it was only tested on 32 signals with 12 descriptions each. In contrast, HapticLLaMA is the first sensory LLM to generate natural language captions for vibrations, highlighting the broad range of human experiences related to sensory, emotional, and associative aspects of vibration perception.

\subsection{Multimodal Captioning}
Multimodal models for image~\cite{vinyals2015show,mokady2021clipcap}, video~\cite{iashin2020multi,wu2023cap4video}, and audio captioning~\cite{zhang2022caption,liu2022visually} have rapidly advanced in the last decade.
Recent advances in open-source LLMs such as LLaMA \cite{DBLP:journals/corr/abs-2302-13971} have accelerated the development of multimodal models. Parameter-Efficient Fine-Tuning (PEFT) \cite{xu2023parameter,xin2024v} adapts pretrained models by updating a small subset of parameters or lightweight modules, significantly reducing computational and storage costs. PEFT has been widely adopted in recent studies \cite{cheng2024emotion,gema2023parameter}. In the field of haptic-language understanding, \citet{hu2025hapticcap} recently introduced HapticCap, a dataset of 92,070 vibration samples paired with sensory, emotional, and associative descriptions, laying the foundation for sensory language models in haptics. In our work, we use HapticCap and PEFT to train the first stage of HapticLLaMA, and in the second stage, we incorporate reinforcement learning \cite{mnih2015human} guided by human feedback.

\section{HapticLLaMA \includegraphics[width=1.35em]{figs/bee.png}}

\subsection{Task Definition}
Similar to image~\cite{vinyals2015show,mokady2021clipcap} and audio captioning~\cite{zhang2022caption,liu2022visually}, haptic captioning involves generating textual descriptions (i.e., captions) from haptic signals. Given a vibration signal $S$ and a target category $c\in \{\text{sensory}, \text{emotional}, \text{associative}\}$, where sensory refers to physical attributes (e.g., intensity of tapping), emotional denotes affective impressions (e.g., the mood of a scene), and associative indicates real-world familiar experiences (e.g., buzzing of a bee, a heartbeat), the goal is to generate a caption corresponding to the specified category of haptic experience.

\subsection{Overall Architecture}
As shown in Figure \ref{fig:architecture}, HapticLLaMA is built on the LLaMA architecture and consists of a haptic tokenizer, a LLaMA model enhanced with LoRA, and a human feedback module trained using reinforcement learning. 
We process the haptic signals offline, converting them into sequences of discrete tokens using two methods: a frequency-based tokenizer using spectral frequency information and an EnCodec-based pretrained neural audio codec \cite{DBLP:journals/tmlr/DefossezCSA23}. The haptic tokens and target category are formatted into a multimodal prompt and input to a LLaMA model fine-tuned with LoRA. Following supervised fine-tuning, human ratings of the generated captions are collected to further refine the model via Direct Preference Optimization  \cite[DPO;][]{rafailov2023direct}. 

\subsection{Input Formalization}
\subsubsection{Haptic Tokenizer}
\begin{figure}[t]
\centering
\subfigure[\label{fig:tokenizer:a}Frequency-based tokenizer.]{
    \includegraphics[width=0.98\linewidth]{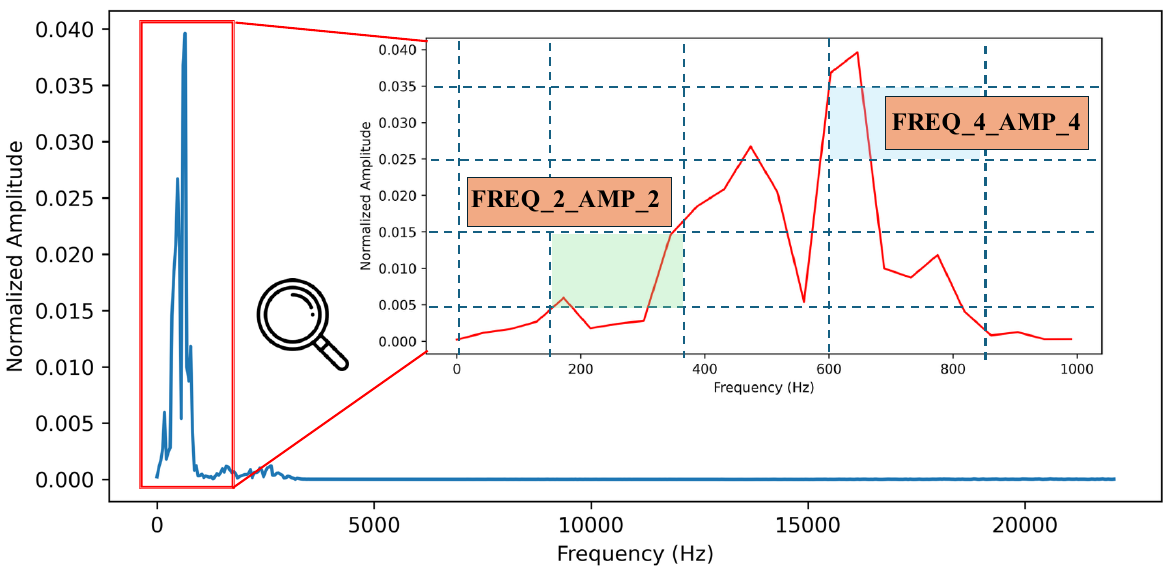}
    \vspace{4pt}
}
\subfigure[\label{fig:tokenizer:b}EnCodec-based tokenizer.]{
    \includegraphics[width=0.97\linewidth]{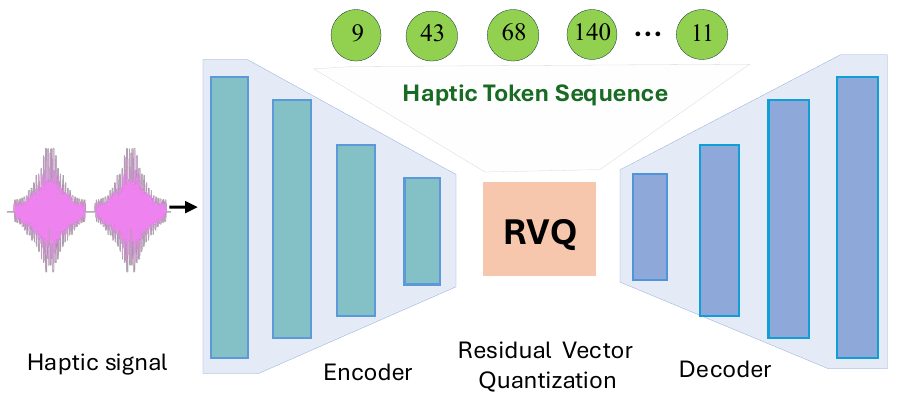}
}
\caption{Two haptic tokenizers to encode vibration signals into input representations suitable for LLaMA.}
\label{fig:haptic_tokenizer}
\end{figure}

\paragraph{Frequency-based Tokenizer:}
This tokenizer (Figure \ref{fig:tokenizer:a}) is motivated by the importance of spectral frequency information in characterizing vibrations \cite{jones2008tactile,bensmaia2005vibrotactile}. Unlike the time domain, where signals are represented by amplitude values over time, the frequency domain represents a signal as a sum of sinusoids at varying frequencies.
The frequency-based tokenizer begins by converting the time-domain signal into the frequency domain via the Fast Fourier Transform (FFT), and subsequently discretizes the resulting frequency components through binning into variable-width intervals. 
Our proposed tokenizer divides the frequency range into logarithmically spaced bins that correspond to just-noticeable differences in human frequency perception \cite{choi2013,israr2006frequency}. Similarly, the amplitude range is segmented into normalized levels. The tokenizer then assigns a unique token (e.g., FREQ\_3\_AMP\_2) to each frequency-amplitude pair, encoding the signal’s spectral content into a form interpretable by LLMs. 

\paragraph{EnCodec-based Tokenizer:}
EnCodec\footnote{\url{https://huggingface.co/facebook/encodec\_24khz}} is a neural audio codec that compresses audio using deep learning \cite{DBLP:journals/tmlr/DefossezCSA23}. It consists of three main components: (1) an encoder that transforms raw audio into a lower-dimensional latent representation, (2) a quantizer that discretizes the latent features via residual vector quantization \cite{zeghidour2021soundstream}, and (3) a decoder that reconstructs the waveform from the quantized codes.
Since vibrations have perceptual and rhythmic similarities to audio signals~\cite{bernard2022rhythm,patzold2023audio,yun2023generating,degraen2021weirding}, we adopt the discrete codes produced by the quantizer as haptic tokens for HapticLLaMA (Figure \ref{fig:tokenizer:b}). 

\subsubsection{Input Format}
After tokenization, the haptic tokens are added to the vocabulary of LLaMA tokenizer and their embeddings are updated in the learning process. 
Given an input tuple $\langle s, c \rangle$, where $s$ represents the signal token sequence generated by the haptic tokenizer and $c$ denotes the category given from $\{\text{sensory}, \text{emotional}, \text{associative}\}$. For the input of HapticLLaMA, we concatenate haptic, category, textual prompt, and special tokens (e.g., <EOS>) as a multimodal prompt $I$. We use the prompt \textit{$I$=``\texttt{haptic signal: <haptic tokens>, its <category> description is: }<caption>.''}. During training, we append the human-written caption as reference.
The auxiliary text tokens (e.g., \textit{``its''}, \textit{``description''}, and \textit{``is''}) can be interpreted as prompts that guide the model’s output.

\subsection{Multimodal HapticLLaMA Model}
Algorithm \ref{algo:hapticllama_training} outlines the two-stage training procedure of HapticLLaMA, consisting of (1) supervised fine-tuning with LoRA adaptation and (2) subsequent fine-tuning based on human feedback on generated captions, as detailed below.

\subsubsection{Stage 1: Supervised Fine-Tuning}
We adopt LLaMA3\footnote{\url{https://huggingface.co/meta-llama/Llama-3.2-3B}} as the backbone of HapticLLaMA. To align the haptic and text modalities, we incorporate haptic tokens into the LLaMA tokenizer's vocabulary by using randomly initialized special tokens in the LLaMA vocabulary. Their embeddings are updated during training, enabling the language model to effectively interpret and utilize the haptic tokens.
For efficient fine-tuning, we employ Low-Rank Adaptation \cite[LoRA;][]{hu2022lora}, which inserts trainable low-rank matrices $\Delta W_*$ into the model weights.
\begin{align}
\begin{split}
     &W_{*}^{\text{lora}} = W_* + \Delta W_*\\
    &= W_* +  B_* A_*, *\in \{Q, V\}
\end{split}
\end{align}
where $W_{*}^{\text{lora}}$ denotes the LoRA-adapted weight matrix, and $W_*$ refers to the original pretrained weight matrix (e.g., the query $Q$ or value $V$ projection weights in a transformer). The subscripts $* \in \{Q, V\}$ indicate the Query and Value projection layers, respectively, and only the matrices $A_*$ and $B_*$ are trainable parameters.

\begin{algorithm}[!t]
\centering
\caption{HapticLLaMA Training}
\begin{algorithmic}[1]
\REQUIRE HapticCap Train set $T=(S,D)$, VibRate signal set $S^{\prime}$, describe category $c$, prompt $p$, LLaMA, haptic tokenizer.
\ENSURE Every signal is associated with a corresponding description.\\
\noindent\textbf{Stage 1: Supervised Fine-Tuning}\\
\STATE Load pretrained LLaMA.
\STATE Tokenize each haptic signal into discrete haptic tokens $S=\{s_{1}, \dots, s_{n}\}$.
\STATE Include haptic tokens and update LLaMA's vocabulary.
\STATE Formalize input based on $S$, $p$, and $c$\\
into $[p_{1}, \cdots,s_{i},\cdots,p_{i},\cdots,\text{<EOS>}]$.
\STATE Supervised fine-tune HapticLLaMA with formalized input.\\
\RETURN initial HapticLLaMA $\mathcal{H}$.\\
\textbf{Stage 2: Fine-Tuning via RLHF}
\STATE Infer the caption $D^{\prime}$ of $s\in S^{\prime}$ using \\
trained $\mathcal{H}$.
\STATE Construct the VibRate dataset by \\ incorporating human ratings $R$.
\STATE Construct caption preference pairs set $\hat{T}$ \\ based on VibRate.
\STATE Fine-tune HapticLLaMA $\mathcal{H}$ with $\hat{T}$.
\RETURN The final HapticLLaMA\includegraphics[height=1.9ex]{figs/bee.png}.
\end{algorithmic}
\label{algo:hapticllama_training}
\end{algorithm}

\subsubsection{Stage 2: Fine-Tuning via RLHF}
Using the trained model from Stage 1, we generate captions for each vibration signal in \textbf{VibRate} dataset (See Section \ref{sec:vib} for details) across sensory, emotional, and associative categories and collect user ratings for the captions on 1-7 Likert scale. The rated captions are then paired into preferred and rejected captions for fine-tuning using the DPO policy\footnote{\url{https://huggingface.co/docs/trl/dpo\_trainer}}. 

\paragraph{Preference Pairs:} 
To create preference pairs for DPO, we match captions rated above the midpoint of the scale with those rated below it. 
Rated captions are split into high-rated (positive) and low-rated (negative) groups using a 3.5 threshold. Each high-rated caption is set as the preferred choice and paired with all low-rated captions as the non-preferred (rejected) ones. 
The HapticLLaMA model is then trained to assign higher likelihood to the preferred response over the rejected one, aligning generation with human preferences.


\begin{table*}[!ht]
\centering
\resizebox{\textwidth}{!}{
\begin{tabular}{lcccc}
\toprule
{\bf Models}                 & {\bf BLEU-1} & {\bf BLEU-4} & {\bf ROUGE-L} & {\bf METEOR} \\
\toprule
Random  &  $5.77$   &  $2.59$  & $10.62$    &   $11.07$    \\
LLaMA (signal-agnostic) & $9.39$ & $2.71$ &$20.67$ & $23.84$   \\
GPT-4.5 (signal-agnostic) &$11.53$& $4.28$ &$26.09$ & $28.20$  \\
Frequency tokens + LLaMA  &  $28.46$    &       $7.48$  & $29.49$     &   $35.24$      \\
AST+GPT-2 &43.51	&16. 47	&35.61	&44.84\\
EnCodec tokens + LLaMA  & $30.07$ & $8.61$                       & $29.50$   &  $35.26$ \\
Frequency tokens + LLaMA + LoRA   &   $40.12 {\scriptstyle \pm 5.1}$   &     $23.16{\scriptstyle \pm 4.3}$                     &  $42.95{\scriptstyle \pm 3.4}$     &   $50.16{\scriptstyle \pm 2.1}$    \\
EnCodec tokens + LLaMA + LoRA  &    $41.54 {\scriptstyle \pm 3.2}$     &    $24.28{\scriptstyle \pm 2.1}$           &   $44.71{\scriptstyle \pm 2.4}$   &   $54.03{\scriptstyle \pm 2.3}$    \\
\midrule
\rowcolor{yellow!10}\textbf{Frequency HapticLLaMA\includegraphics[width=1.35em]{figs/bee.png}}: Frequency tokens + LLaMA + LoRA + RLHF &  $49.43{\scriptstyle \pm 1.4}$          &   $ 30.86{\scriptstyle \pm 1.5}$        & $ 48.74{\scriptstyle \pm 2.1}$             &    $58.95{\scriptstyle \pm 2.1}$               \\
\rowcolor{green!10}\textbf{EnCodec HapticLLaMA\includegraphics[width=1.35em]{figs/bee.png}}: EnCodec tokens + LLaMA + LoRA + RLHF& $51.36{\scriptstyle \pm 1.3}$             &  $ 32.06{\scriptstyle \pm 1.2}$             &   $49.56{\scriptstyle \pm 1.6}$    &    $59.98{\scriptstyle \pm 1.8}$    \\

\bottomrule
\end{tabular}}
\caption{Results on automated n-gram metrics. Baselines include Random, LLaMA, and GPT-4.5. Ablation results show the impact of each component and the performance of HapticLLaMA with Frequency and EnCodec tokens. Green and yellow denote the best and second-best performances, respectively. $\pm$ denotes standard deviation.}
\label{tab:main_results}
\end{table*}

\section{VibRate Dataset Construction}
\label{sec:vib}
VibRate is a diverse, manually curated dataset of 16,896 <vibration, caption, rating> tuples (see Appendix \ref{sec:compensation} for details on recruitment and payment).
It includes 704 vibrations constructed from four diverse sources: (a) 174 vibration signals are created by varying signal parameters (e.g., frequency); (b) 180 vibrations derived from sound effect libraries by mimicking timing or applying low-pass filtering \cite{ternes2008designing, degraen2021weirding, yun2023generating}; (c) 176 vibrations generated by HapticGen \cite{sung2025hapticgen}; and (d) 174 custom-made vibrations created manually through signal transformations such as time reversal, repetition, and mixing \cite{schneider2016studying, maclean_multisensory_2017}.
For each signal, we generate four captions in each category: two using frequency-based $\mathcal{H}$ and two from EnCodec-based $\mathcal{H}$. Then, we collect ratings from 44 human evaluators for the generated captions.
Annotators are instructed to assign a final rating on a scale of 1 (poor) to 7 (excellent) based on two criteria: (1) the clarity and semantic accuracy of the caption, and (2) the alignment between the haptic vibration experience and caption. A higher rating indicates better quality and closer alignment with human perception. In total, 44 users participated, each evaluating captions for 32 different vibrations, yielding 16,896 <vibration, caption, human rating> samples.

\section{Experiments}


\paragraph{Datasets:} 
We use two datasets to develop and evaluate HapticLLaMA: \textbf{HapticCap} \cite{hu2025hapticcap} and \textbf{VibRate} on two stages respectively. \textbf{HapticCap} is the largest fully human-annotated haptic-captioned dataset, containing 92,070 haptic-text pairs, with 8-10 user-written captions per vibration, describing sensory, emotional, and associative attributes. 
The HapticCap dataset is divided into training, validation, and test sets (see Appendix \ref{sec:data_split}). We evaluate HapticLLaMA from both Stage 1 and Stage 2 on the test set.

\paragraph{Baselines:}
We set the following baselines: (1) \textbf{Random}: For each vibration in the test set, we randomly select one caption from the candidate set as the caption for that signal, (2) \textbf{Signal-agnostic}: 
Since LLMs can generate fluent descriptions grounded in their pretraining and guided through prompting, we examine the capability of LLaMA3.2-3B and GPT-4.5 \footnote{\url{https://platform.openai.com/docs/models/gpt-4.5-preview}} in generating captions without receiving any signal in the input (see Appendix \ref{sec:prompt} for details), (3) AST + GPT-2: AST serves as the HapticEncoder (analogous to the CLIP encoder in ClipCap) to encode raw haptic signal into haptic features, and GPT-2 is used as the caption decoder (i.e., non-LLM sequence model) for generating haptic descriptions.
(4) \textbf{Without LoRA Finetuning}: Fine-tuning is disabled by removing LoRA from the Frequency-based and EnCodec-based HapticLLaMA models, while keeping haptic token training.
(5) \textbf{Without RLHF}: Model performance with and without DPO fine-tuning from human feedback is reported.

\paragraph{Evaluation Metrics:}
Following prior work on captioning \cite{drossos2020clotho,cui2018learning}, we evaluate haptic captions using BLEU-1 \cite{papineni2002bleu}, BLEU-4 \cite{papineni2002bleu}, ROUGE-L \cite{lin2004rouge}, and METEOR \cite{banerjee2005meteor} metrics. Due to the inherent ambiguity of haptic perception, each vibration in HapticCap is associated with 8–10 ground-truth reference captions. We evaluate each prediction using the reference captions and report the mean and standard deviation of the metrics to reflect performance and variability. Human evaluations are the primary benchmark, while automated metrics offer a complementary but imperfect proxy for scalable evaluation and ablation.
Implementation details are provided in Appendix \ref{sec:implementation}.

\subsection{Performance on Automated Metrics}
Table \ref{tab:main_results} presents the performance of our Frequency HapticLLaMA and EnCodec HapticLLaMA, the baselines, as well as the ablation results on HapticLLaMA with different haptic tokenizers, LoRA adapter, and RLHF.

The random baseline exhibits consistently poor performance across all evaluation metrics. In contrast, the signal-agnostic LLaMA and GPT-4.5 models achieve marginal improvements, attributable to their advanced language generation capabilities despite the absence of haptic signals. 
Ablation results demonstrate the contribution of each component in HapticLLaMA. Incorporating haptic tokens into the training vocabulary leads to consistent improvements across all evaluation metrics. LoRA fine-tuning with haptic tokenizers shows strong absolute gains between +11.47 (BLEU-1) to +18.77 (METEOR) points in all metrics. Applying RLHF via DPO further boosts HapticLLaMA's performance across all metrics, including an absolute gain of +7.78 on BLEU-4 score.



EnCodec tokens consistently outperform Frequency tokens by a small margin, with EnCodec HapticLLaMA (last row) achieving the highest scores across all metrics. This result may be because the Frequency-based tokenizer emphasizes the spectral domain while neglecting temporal and rhythmic features of vibration, whereas the EnCodec-based tokenizer captures richer representations of temporal changes based on the raw vibration signal and creates efficient tokens for signal compression and reconstruction. Additionally, the length of haptic token sequences and vocabulary size produced by the EnCodec tokenizer are significantly higher than that of the Frequency-based tokenizer (see Section~\ref{sec:haptic_token}), suggesting that EnCodec captures more fine-grained variations in the vibration signals. These results demonstrate that both frequency and EnCodec tokens can capture input signals, and the application of DPO further enhances caption generation quality by refining the alignment between model output and human preferences.


\begin{figure}[!t]
\centering
\subfigure{\label{fig:subfig:b}}\addtocounter{subfigure}{-1}
\subfigure[After Stage 1.]
{\includegraphics[width=0.475\linewidth]{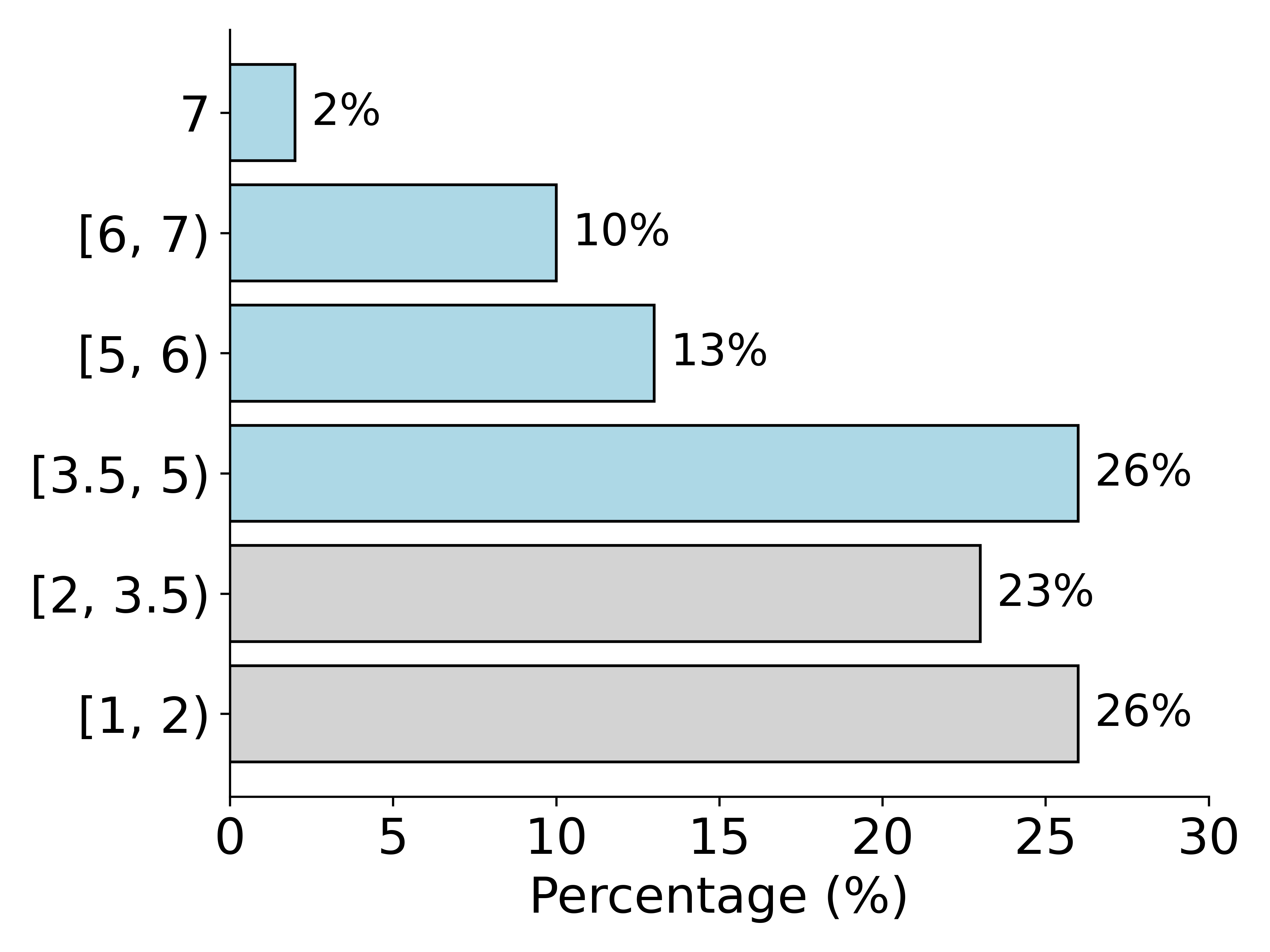}}
\subfigure{\label{fig:subfig:a}}\addtocounter{subfigure}{-1} 
\subfigure[After Stage 2.]
{\includegraphics[width=0.475\linewidth]{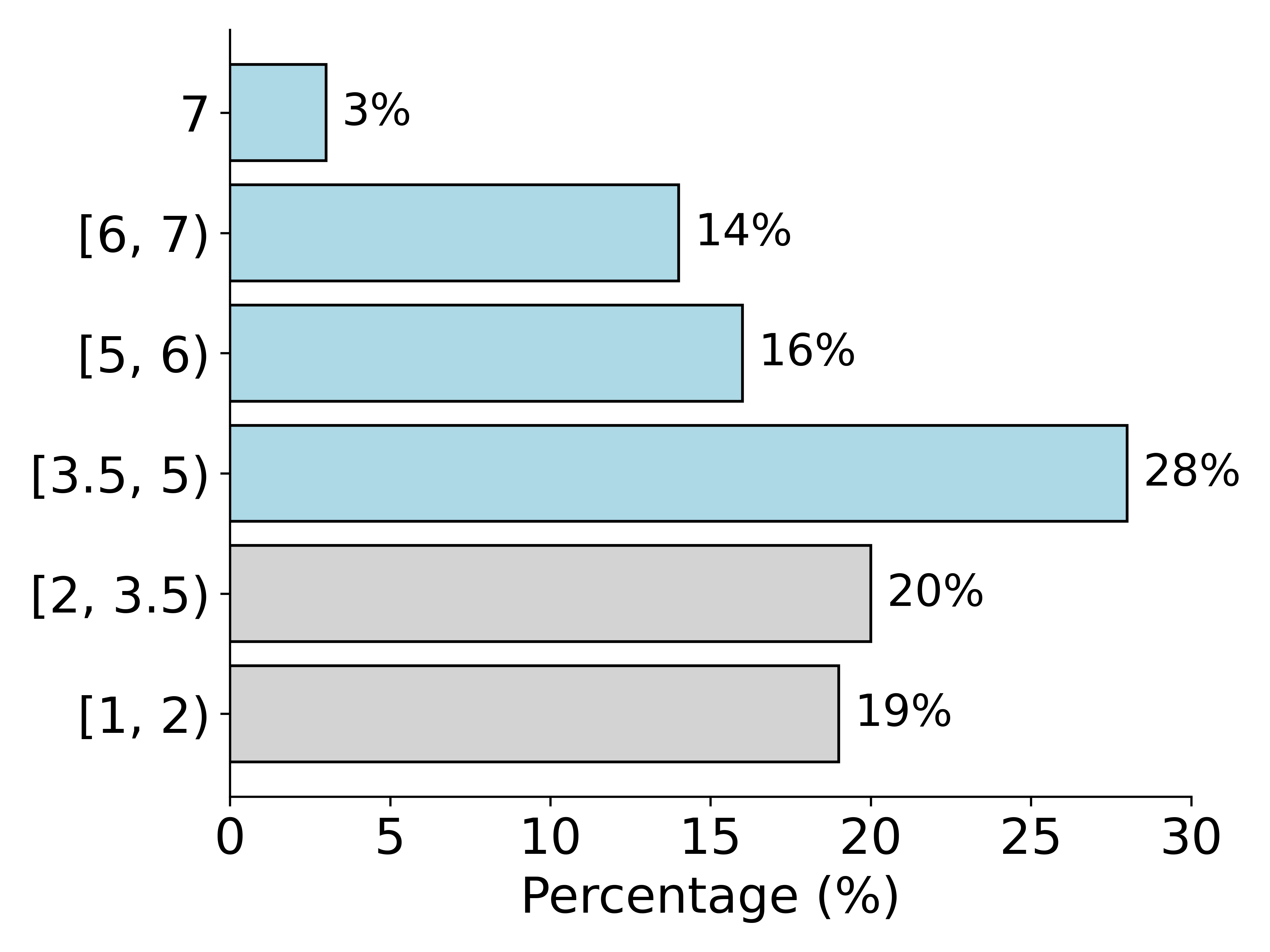}}
\caption{The distribution of human ratings after Stage 1 and 2 training for captions generated by EnCodec-based HapticLLaMA. 
Higher ratings are shown in blue and lower ratings are in gray. }
\label{fig:heatmap_bar}
\end{figure}

\subsection{Human Evaluation}
Figure \ref{fig:heatmap_bar} presents the human evaluation results of descriptions generated by HapticLLaMA in Stage 1 and Stage 2, respectively.
We use the human ratings from VibRate as the evaluation results after Stage 1.
After completing both training stages, we randomly select 20 vibration signals from the HapticCap test set and generate corresponding haptic captions using the final HapticLLaMA model. Two evaluators are instructed to provide final ratings on a 1–7 scale, using the same human rating setup as in Stage 1. The results show a noticeable improvement in the quality of generated captions after Stage 2, with over 61\% of ratings above 3.5. 
Specifically, there is a visible shift in the distribution toward higher rating intervals, indicating that more captions are perceived as relevant and accurate after Stage 2. For instance, the proportion of higher ratings increases across all intervals after Stage 2, with about 10\% increase in ratings above 3.5 and a particularly notable rise of about 3\% and 4\% in the range [5, 6) and [6, 7), respectively. In contrast, the share of lower ratings, such as [1, 2) and [2, 3.5), decreases significantly by around 7\% and 3\%, respectively. These results indicate that the preference alignment introduced in Stage 2 notably improves the model’s ability to generate descriptions that are semantically clearer and perceptually aligned with human judgments.

\begin{figure}[!t]
\centering
\subfigure[BLEU-4 scores.\label{fig:visualization:a}]{
\includegraphics[width=0.9\linewidth]{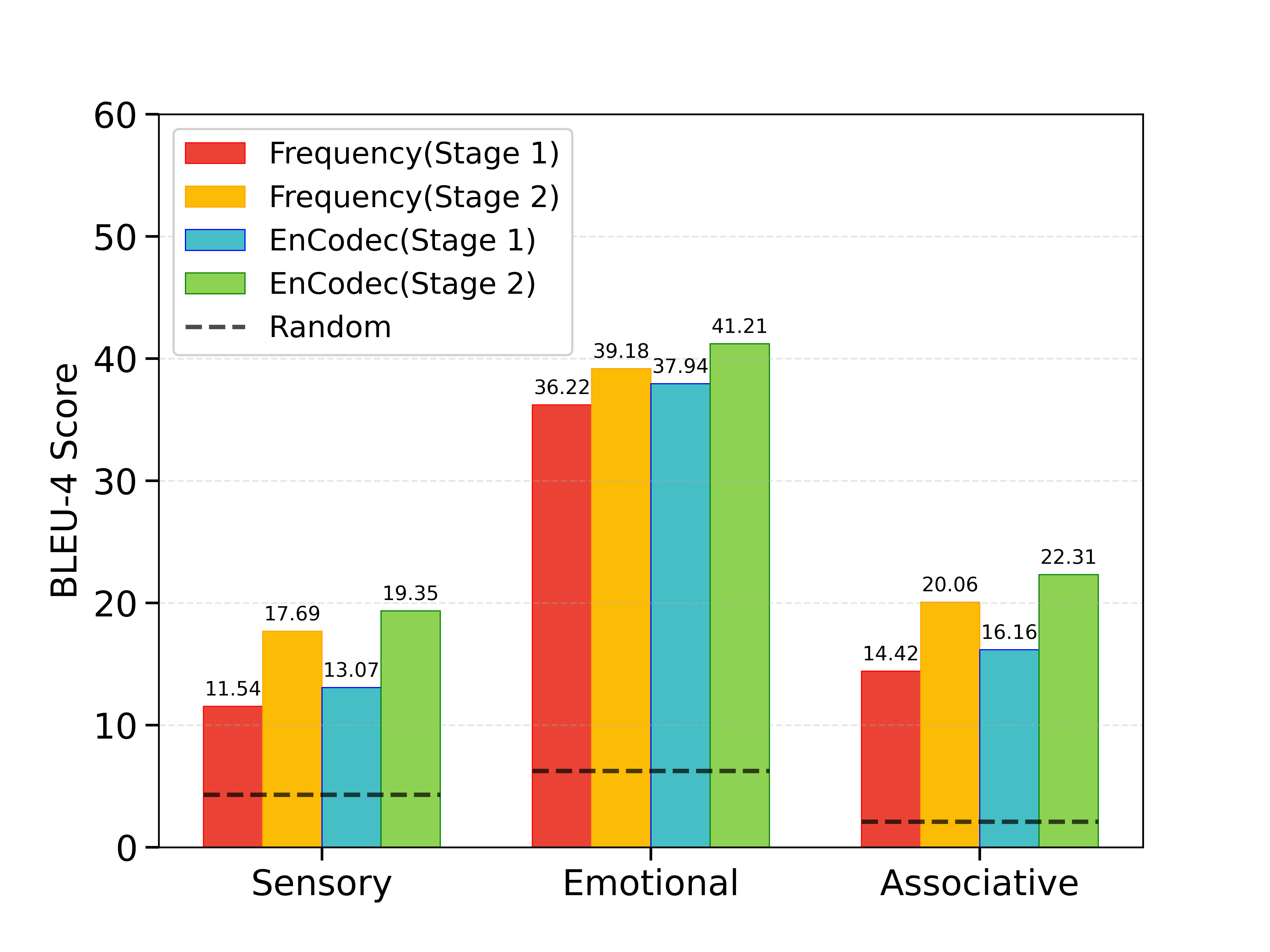}}
\hspace{0.05in}
\subfigure[Average human ratings.\label{fig:visualization:b}]{
\includegraphics[width=0.8\linewidth]{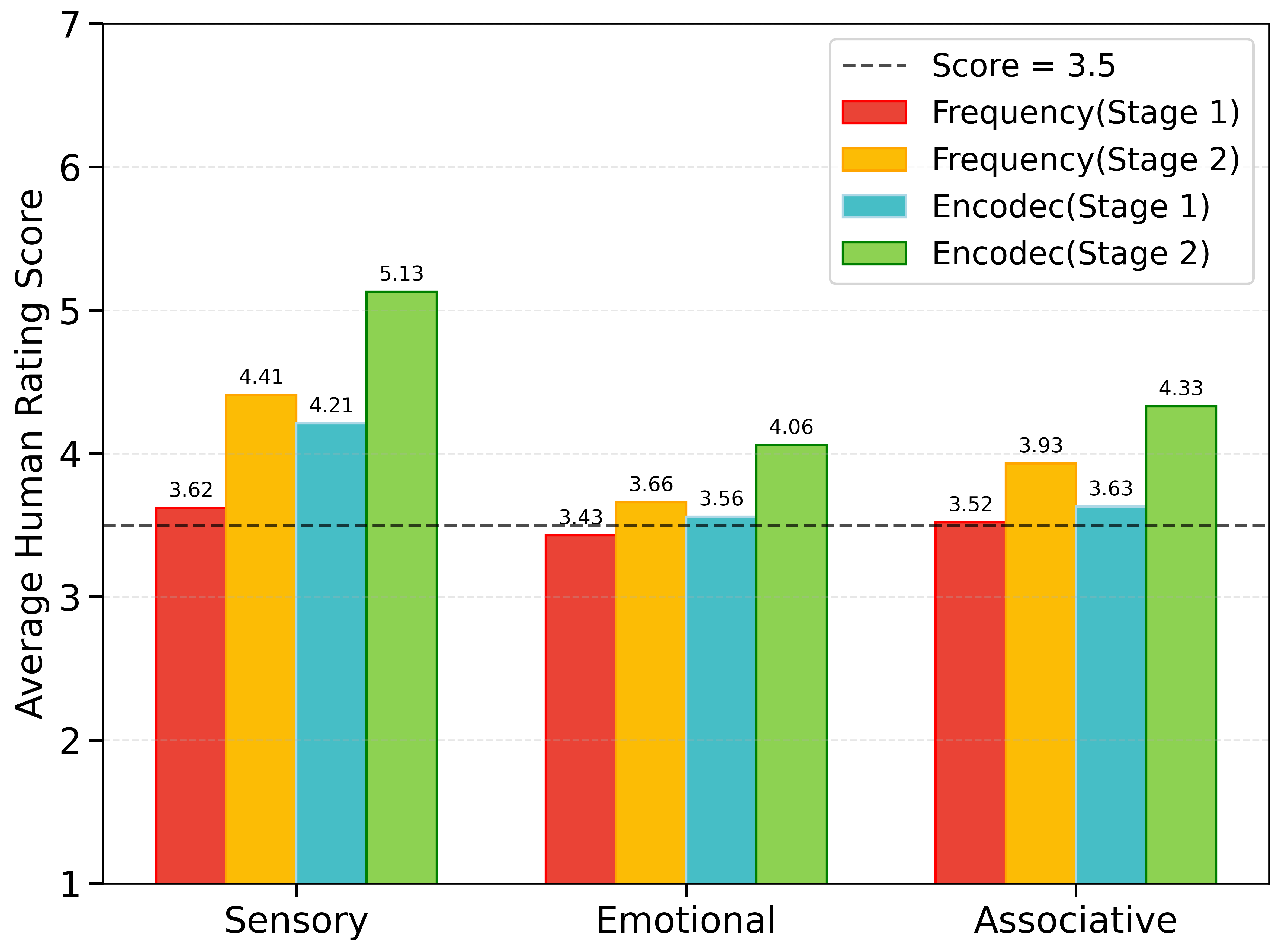}}
\caption{EnCodec-based HapticLLaMA's performance across sensory, emotional, and associative categories.}
\label{fig:category_performance}
\end{figure}

\subsection{Category-Specific Performance}
Figure \ref{fig:category_performance} compares HapticLLaMA variants across sensory, emotional, and associative categories using (a) BLEU-4 scores (with additional METEOR results in Appendix \ref{sec:more_results}) and (b) average human ratings. Both illustrate that the application of DPO consistently improves performance in all categories, yielding gains in BLEU-4 and average human rating, respectively, in line with the observed improvements in overall performance.

As measured by BLEU-4, emotional captions yield the highest performance among the three categories across both HapticLLaMA variants, with the EnCodec-based model achieving the leading score of 41.2 in BLEU-4. This outcome may be attributed to the fact that emotional captions tend to show higher agreement among multiple human annotators \cite{hu2025hapticcap}, thereby facilitating more accurate and coherent generation. According to human evaluations, sensory captions receive the highest ratings across all HapticLLaMA variants and stages. The discrepancy between automatic metrics and human judgments may stem from the limitations of automated methods (e.g., BLEU), which focus on surface-level textual similarity, whereas human raters are more attuned to the qualitative aspects of haptic experience. 
Similar discrepancy between automated metrics and human judgments are reported in image captioning literature \cite{elliott2014comparing,kilickaya2016re}.

\begin{figure*}[!ht]
\centerline{\includegraphics[width=\textwidth]{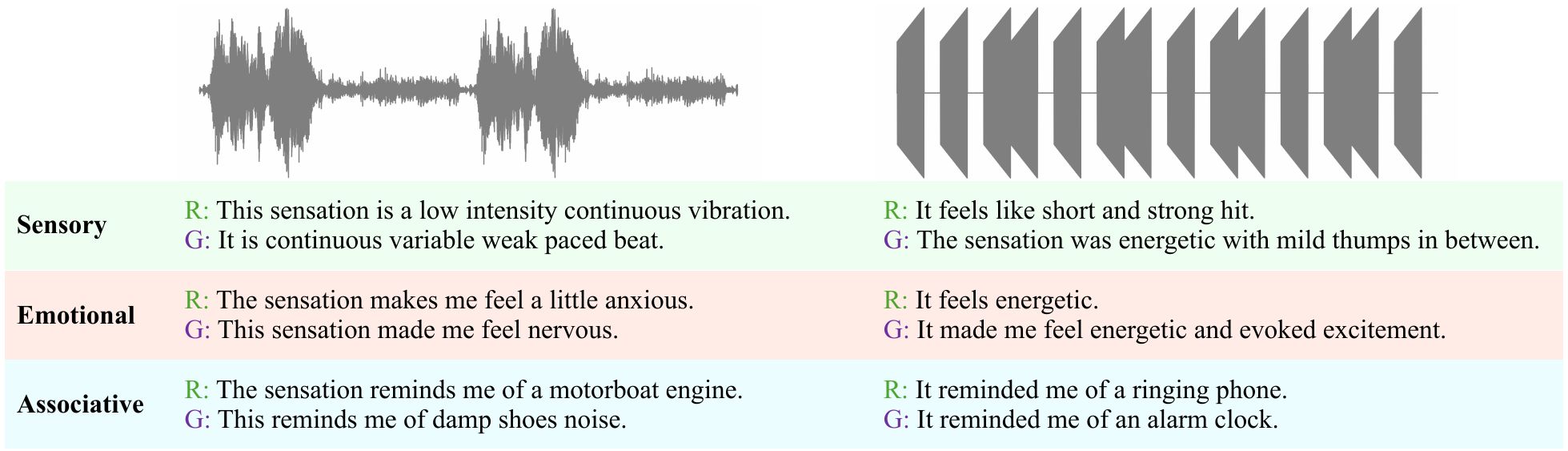}}
\caption{Case study showing two example vibrations with captions. ``R'' (in green) denotes reference ground-truth captions from HapticCap, while ``G'' (in purple) denotes captions generated by EnCodec-based HapticLLaMA.}
\label{fig:case_study}
\end{figure*}



\subsection{Case Study}
Figure~\ref{fig:case_study} illustrates a comparison between reference ground-truth (R) and generated (G) captions for two haptic signals from the test set.
For the left signal, we observe a continuous haptic vibration featuring two prominent vibration peaks in the middle. Compared with the reference captions, the generated ones effectively capture continuity, weak intensity, and rhythm of the vibration in the sensory category. 
For the emotion dimension, the generated caption also successfully reflect emotional responses (e.g., ``nervous'' $\leftrightarrow$ ``anxious''). 
Regarding the associative category, HapticLLaMA broadens the range of associations from ``motorboat engine'' to ``damp shoes'', both capturing dull, heavy sensations and demonstrating a broader and more diverse associative space. The second signal displays features that are clearly distinct from those of the first. The vibration is an intermittent sequence of pulses with pauses at regular intervals. In the sensory caption, HapticLLaMA captures the pulsing characteristic (``short hit'' $\leftrightarrow$ ``mild thumps in between'') and overall energy (``strong'' $\leftrightarrow$ ``energetic'').
For the emotional aspect, HapticLLaMA aligns with the overall sentiment and enhances it (e.g., adding ``evoked excitement''). In terms of associative categories, ``ringing phone'' and ``alarm clock'' sensations are logically similar, with both involving sequences of strong pulses in mobile phones.

\subsection{Haptic Tokenizer Analysis}
\label{sec:haptic_token}

Table \ref{tab:haptic_token} presents the summary statistics of haptic tokens for two tokenizers: Frequency and EnCodec. EnCodec-based tokenizer has a significantly larger haptic token vocabulary (1,024) compared to the Frequency-based tokenizer (278). This suggests that EnCodec can represent haptic signals with greater granularity and expressive power, consistent with the improved performance observed in the EnCodec-based HapticLLaMA. The Frequency-based tokenizer produces shorter sequences when representing a vibration signal, with an average length of 47.5 tokens, indicating a more compact representation. In contrast, EnCodec outputs a fixed-length sequence of 1,379 tokens for all signals, which may introduce redundancy but ensures uniform input sizes for downstream models. In summary, the Frequency-based tokenizer offers higher compression efficiency, making it suitable for lightweight or real-time applications but slightly inferior performance. On the other hand, EnCodec provides more detailed and consistent representations, which may be beneficial for tasks requiring richer signal understanding.

\begin{table}[!t]
\centering
\resizebox{0.87\linewidth}{!}{
\begin{tabular}{llc}
\toprule
{\bf Tokenizer}                 & {\bf Item}                        & {\bf Count} \\
\toprule
\multirow{4}{*}{Frequency} & Haptic token vocabulary size  & 278     \\
                           & Average haptic signal length & 47.5    \\
                           & Max haptic signal length     & 52     \\
                           & Min haptic signal length     & 12     \\
\midrule
\multirow{4}{*}{EnCodec}   & Haptic token vocabulary size  & 1,024    \\
                           & Average haptic signal length & 1,379    \\
                           & Max haptic signal length     & 1,379     \\
                           & Min haptic signal length     & 1,379 \\  
\bottomrule
\end{tabular}}
\caption{Summary statistics of haptic tokenizers.}
\label{tab:haptic_token}
\end{table}

\section*{Conclusion}
We propose HapticLLaMA, the first multimodal sensory language model trained with two distinct haptic tokenizers for haptic captioning task. Built on the LLaMA architecture, HapticLLaMA is trained in two stages: (1) supervised fine-tuning on the LLaMA architecture with LoRA, and (2) fine-tuning through reinforcement learning from human feedback (RLHF). Our evaluation shows that HapticLLaMA can effectively perceive haptic vibration signals and provide substantial improvements over existing LLMs on the haptic captioning task. Our work advances haptic-language understanding by enabling LLMs to interpret physical sensory data. 

\section*{Limitations}
BLEU, ROUGE, and METEOR metrics are suboptimal for evaluating haptic captioning quality, as they primarily emphasize textual fluency and lexical overlap, while failing to account for the semantic alignment between generated captions and the underlying haptic signals. Human evaluation, although more reliable, is resource-intensive and typically conducted on a small, randomly sampled subset due to high labor costs. As a result, neither automatic nor manual evaluation methods provide a fully accurate assessment of captioning quality. HapticLLaMA focuses on vibration signals as the most accessible and diverse form of haptics, but cannot interpret other forms of haptics, such as force feedback or temperature signals.  

 
\section*{Ethics Statement}
While the idea of HapticLLaMA interpreting haptic vibration signals is appealing and holds potential for applications in human-computer interaction (HCI) and robotics, its current performance remains limited. 
The model achieves only around 32.06\% on BLEU-4, an average human rating of 4.8 on a 7-point scale, and about 61\% of captions rated above 3.5, indicating that its performance remains insufficient for real-world deployment.

\section*{Acknowledgement}
We sincerely thank the volunteers for their generous contributions and invaluable efforts in providing high-quality data annotation, which has been instrumental in supporting our research. This work was supported by research grants from VILLUM FONDEN (VIL50296) and the National Science Foundation (\#2339707).

\bibliography{acl_latex}

\newpage

\appendix

\section{Prompt for Signal-agnostic Experiments}
\label{sec:prompt}
In the absence of haptic signal input (signal-agnostic baseline), LLaMA is trained and tested purely on prompts \textit{``its <category> description is: <caption>''} to evaluate its signal-agnostic capability. We provide the caption during the training and remove it from the prompt during inference. The prompts for GPT-4.5 are designed to guide the model in generating relevant haptic descriptions for sensory, emotional, and associative aspects (see Table \ref{tab:prompt}), following the data collection guidelines in prior work \cite{hu2025hapticcap}. We provide several caption demonstrations in the prompt to guide the generation of GPT-4.5, as shown in Table \ref{tab:example}.

\begin{table}[!h]
\centering
\resizebox{\linewidth}{!}{\begin{tabular}{l|c|ll}
\toprule
                             & {\bf Role}                    & {\bf Prompt}                                    &  \\
\toprule
\multirow{2}{*}{}            & \multirow{2}{*}{system} & You are experiencing a tactile sensation. \\
                             &                         & Describe what you feel in one sentence.   &  \\
Sensory                      & user                    & How would you describe the sensation?     &  \\
\multirow{2}{*}{Emotional}     & \multirow{2}{*}{user}   & How does this sensation make you feel,    &  \\
                             &                         & can you attach any emotion to it?          &  \\
\multirow{2}{*}{Associative} & \multirow{2}{*}{user}   & Can you associate any action              &  \\
                             &                         & or object with this sensation?             & \\
\bottomrule
\end{tabular}}
\caption{The prompt templates for GPT-4.5. }
\label{tab:prompt}
\end{table}

\begin{table}[!ht]
\centering
\resizebox{\linewidth}{!}{\begin{tabular}{l|c|ll}
\toprule
                             & {\bf Role}                       & {\bf Response}                            &  \\
\toprule
\multirow{2}{*}{Sensory}     & \multirow{2}{*}{assistant} & sensory: It is a steady, regular and strong &  \\
                             &                            &pulsation   &  \\
\multirow{2}{*}{Emotional}   & \multirow{2}{*}{assistant} & emotional: It makes me feel calm, peaceful      &  \\
                             &                            & and focused.      &  \\
\multirow{1}{*}{Associative} & \multirow{1}{*}{assistant} & associative: It's quite like a machine gun.     &  \\
\bottomrule
\end{tabular}}
\caption{The demonstrations for GPT-4.5. }
\label{tab:example}
\end{table}

\section{Details of Data Split}
\label{sec:data_split}
Table \ref{tab:data_split} presents the data split of the HapticCap dataset, along with the category distributions across sensory, emotional, and associative captions.
\begin{table}[!h]
\centering
\resizebox{0.85\linewidth}{!}{\begin{tabular}{l|ccc}
\toprule
 \bf Category    & \bf Train &\bf Valid &\bf Test\\
\toprule
Sensory     & 24,641  & 2,677   & 3,372  \\
Emotional    & 24,641     & 2,677      &  3,372  \\
Associative & 24,641  & 2,677   & 3,372  \\
All & 73,923  & 8,031   & 10,116 \\
\bottomrule
\end{tabular}}
\caption{The details of HapticCap. }
\label{tab:data_split}
\end{table}

\section{Implementation Details}
\label{sec:implementation}
All experiments are conducted on NVIDIA RTX A100 and RTX H100 GPUs. Due to limited computational resources, we adopt LLaMA3.2-3B 
as the backbone of HapticLLaMA, and integrate Low-Rank Adaptation (LoRA) into all query and value projection layers of the Transformer architecture to enable parameter-efficient fine-tuning. In Stage 1, the batch size is set to 4, with an overall learning rate of 3e-4. The model is trained with a generative loss. For the frequency-based tokenizer, the bin edges are constructed using geometric spacing (via np.geomspace) based on the minimum and maximum values of each signal.

In Stage 2, we adopt the default setting of DPO trainer to optimize HapticLLaMA. The model is optimized the policy to align with human preferences through a simple classification objective, thereby improving caption quality. To construct preference pairs from the VibRate ratings, we proceed as follows: (1) Collect human-rated captions. Each caption generated by the Stage-1 model receives a human rating on a 1–7 scale. (2) Split captions into positive and negative groups. We use a 3.5 threshold (the midpoint of the scale) to separate captions into:high-rated (positive) group: ratings $\geq$ 3.5 and low-rated (negative) group: ratings < 3.5. (3) Form preference pairs. 
For DPO, every caption in the high-rated group is treated as the preferred response. Each preferred caption is then paired with all captions in the low-rated group, which serve as the rejected responses. This creates a set of (preferred, rejected) preference pairs. Regarding the pairing strategy, pairing only low-rated captions would train the model merely to choose the ``less bad'' option, rather than to prefer truly good captions. In contrast, pairing high-rated (positive) with low-rated (negative) captions explicitly encourages the model to favor higher-quality outputs, providing a much stronger and more meaningful supervisory signal.

This results in 2,585 preference pairs: 1,280 sensory, 849 emotional, and 456 associative. Each vibration has four rated captions per description type, allowing up to four preference pairs per category. This setup prevents data explosion during DPO training.

In training with DPO objective, HapticLLaMA learns to assign: (1) higher likelihood to the preferred caption, and (2) lower likelihood to the rejected caption conditioned on the same haptic input. Two models are involved: the policy model, which is trained to learn human preferences with VibRate dataset, and the reference model, which is the supervised fine-tuned model (e.g., the first-stage SFT model) used as the baseline prior to preference optimization. We set $\beta=0.1$ for DPO training.

\begin{figure*}[!ht]
\centerline{\includegraphics[width=\textwidth]{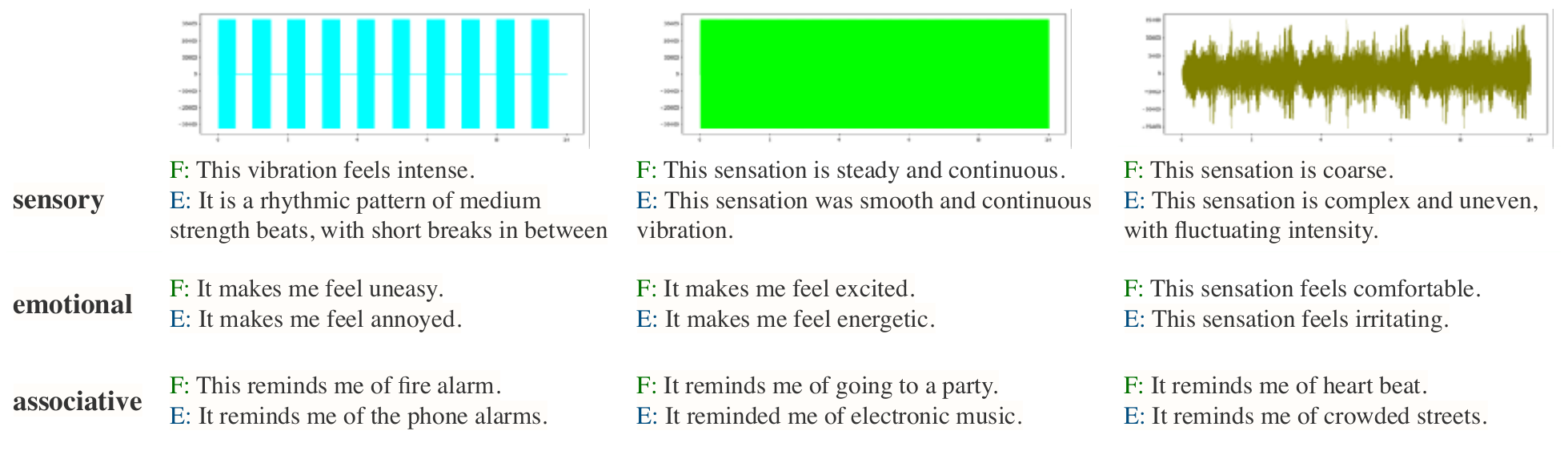}}
\caption{Token-level analysis of three cases (from left to right: Case 1, Case 2, and Case 3), where ``F'' denotes the caption generated by frequency-based HapticLLaMA and ``E'' denotes the caption generated by EnCodec-based HapticLLaMA.}
\label{fig:tokenizer_analysis}
\end{figure*}

\section{More Experimental Results}
\label{sec:more_results}

\subsection{More Performance of HapticLLaMA}
\begin{figure}[!h]
\centerline{\includegraphics[width=0.87\linewidth]{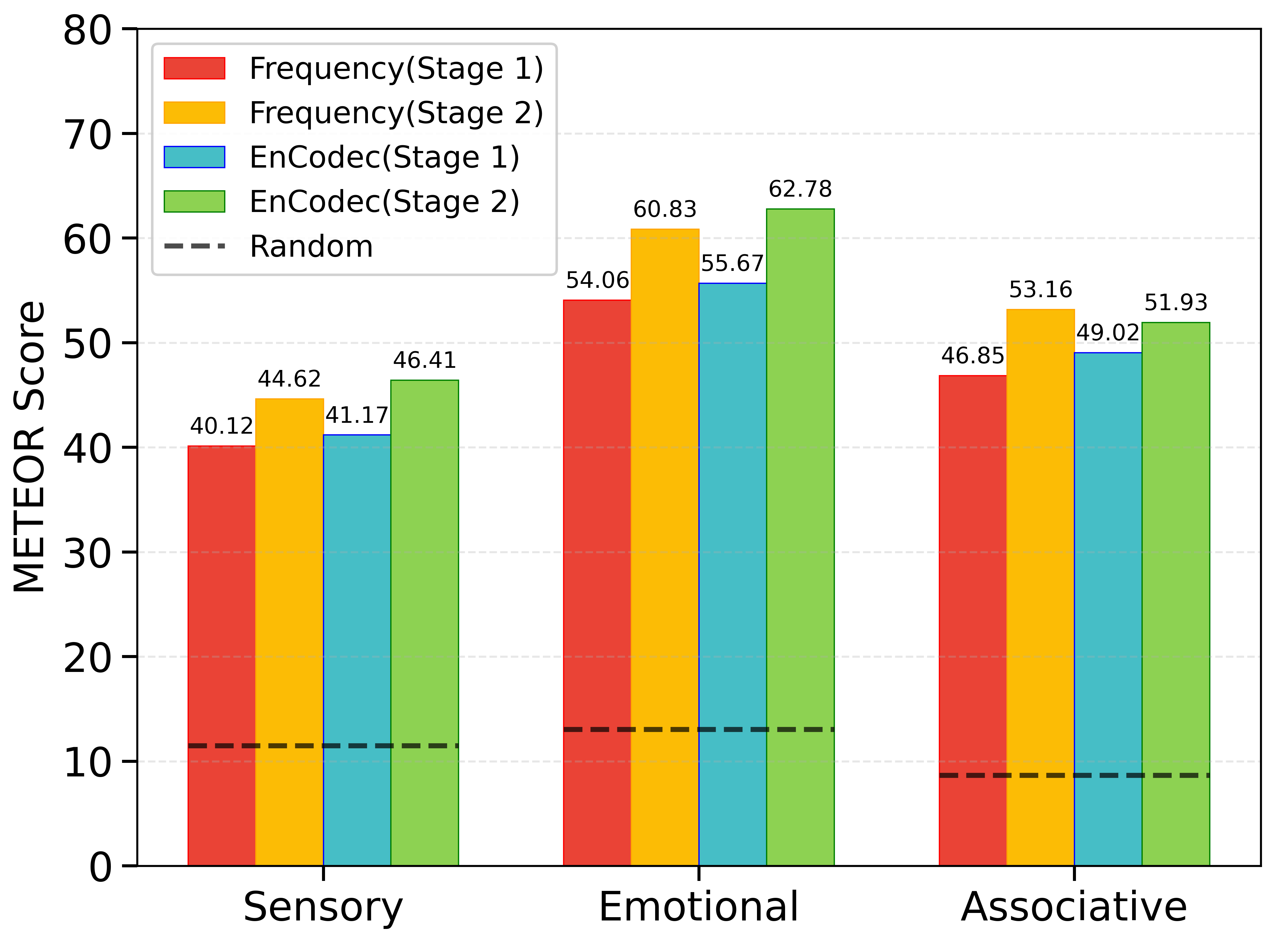}}
\caption{Performance of EnCodec-based HapticLLaMA across sensory, emotional, and associative categories, as measured by METEOR scores.}
\label{fig:meteor}
\end{figure}

Figure \ref{fig:meteor} presents the METEOR scores of EnCodec HapticLLaMA across the three categories. The trends across the three categories are consistent with those observed in BLEU (see Figure \ref{fig:category_performance}(a)).
We also ran a separate experiment and constructed preference pairs with a different approach by taking any two captions associated with the same signal , labeling the higher-rated caption as the accepted choice and the lower-rated one as the rejected choice (max six pairs per signal and description category). This creates a perfectly balanced dataset for each category, butleads to a noticeable performance drop on each category, as shown in Table \ref{tab:sep_category}.
\begin{table}[t]
\centering
\begin{tabular}{lcc}
\hline
\textbf{Category} & \textbf{BLEU-4} & \textbf{METEOR} \\
\hline
Sensory      & 17.81\,($\downarrow$1.54) & 43.16\,($\downarrow$3.25) \\
Emotional    & 38.98\,($\downarrow$2.23) & 58.62\,($\downarrow$4.16) \\
Associative  & 21.04\,($\downarrow$1.27) & 49.46\,($\downarrow$2.47) \\
\hline
\end{tabular}
\caption{Performance comparison in terms of BLEU-4 and METEOR across different categories.}
\label{tab:sep_category}
\end{table}
\subsection{Tokenizer Analysis}
We create three example signals: (1) a regular rhythmic pulse, (2) a continuous signal with a single frequency, and (3) a continuous signal with various frequencies ranging from 50Hz-500Hz. We then generated sensory, emotional, and associative captions using both Frequency HapticLLaMA and EnCodec HapticLLaMA, as shown in Figure \ref{fig:tokenizer_analysis}.

For the rhythmic pulse (case 1), EnCodec HapticLLaMA reliably produced captions that explicitly mentioned temporal rhythm (e.g., ``rhythmic pattern''), whereas Frequency HapticLLaMA tended to describe only overall intensity without capturing rhythmic structure. For the single-frequency signal (case 2), both models produced similar descriptions, indicating that the frequency-based tokenizer is sufficient when no complex temporal dynamics are present. For the multi-frequency signal (case 3), EnCodec HapticLLaMA more often referred to changing or sweeping vibrations, suggesting that the EnCodec tokenizer better captures time-varying spectral patterns.
\section{Data Collection and Compensation}
\label{sec:compensation}
In this work, the participants were university students recruited via advertisements and snowball sampling. Each participant rated captions for 32 signals in a one-hour session and received \$15 USD in cash, which exceeds the local minimum wage. 


\end{document}